\documentclass[sigconf]{acmart}

\usepackage{array}
\usepackage{geometry}
\usepackage{graphicx}
\usepackage{amsmath}
\usepackage{multirow}
\usepackage{arydshln}
\usepackage{caption}
\usepackage{subcaption}
\usepackage{balance}

\usepackage[capitalize]{cleveref}
\crefname{section}{Sec.}{Secs.}
\Crefname{section}{Section}{Sections}
\Crefname{table}{Table}{Tables}
\crefname{table}{Tab.}{Tabs.}

\newcommand{\RomanNumeralCaps}[1]{\uppercase\expandafter{\romannumeral#1}}
\include{pmbox}
\DeclareUnicodeCharacter{251C}{\mbox{\kern.23em
  \vrule height2.2exdepth1exwidth.4pt\vrule height2.2ptdepth-1.8ptwidth.23em}}
\DeclareUnicodeCharacter{2500}{\mbox{\vrule height2.2ptdepth-1.8ptwidth.5em}}
\DeclareUnicodeCharacter{2514}{\mbox{\kern.23em
  \vrule height2.2exdepth-1.8ptwidth.4pt\vrule height2.2ptdepth-1.8ptwidth.23em}}
\DeclareUnicodeCharacter{2502}{\mbox{\kern.23em
  \vrule height2.2exdepth1exwidth.4pt}}
\usepackage{xspace}

\makeatletter
\DeclareRobustCommand\onedot{\futurelet\@let@token\@onedot}
\def\@onedot{\ifx\@let@token.\else.\null\fi\xspace}

\def\ie{\emph{i.e}\onedot}

\newcommand{\MetricName}{Preference Score}
\newcommand{\MetricNameAbbr}{PS}

\AtBeginDocument{%
  }

\copyrightyear{2023}
\acmYear{2023}
\setcopyright{acmlicensed}\acmConference[MM '23]{Proceedings of the 31st ACM International Conference on Multimedia}{October 29-November 3, 2023}{Ottawa, ON, Canada}
\acmBooktitle{Proceedings of the 31st ACM International Conference on Multimedia (MM '23), October 29-November 3, 2023, Ottawa, ON, Canada}
\acmPrice{15.00}
\acmDOI{10.1145/3581783.3612831}
\acmISBN{979-8-4007-0108-5/23/10}
\begin{document}

\title[Learning and Evaluating Human Preferences for Conversational Head Generation]{Learning and Evaluating Human Preferences for\\Conversational Head Generation}

\author{Mohan Zhou}
\orcid{0000-0003-3250-4978}
\affiliation{%
  \institution{Harbin Institute of Technology}
  \country{}
}
\email{mhzhou99@outlook.com}

\author{Yalong Bai}
\orcid{0000-0002-8416-9027}
\affiliation{%
  \institution{JD Explore Academy}
  \country{}
}
\email{ylbai@outlook.com}

\author{Wei Zhang}
\orcid{0000-0002-1492-8286}
\affiliation{%
  \institution{Gaoding AI}
  \country{}
}
\email{wzhang.cu@gmail.com}

\author{Ting Yao}
\orcid{0000-0001-7587-101X}
\affiliation{%
 \institution{HiDream.ai}
 \country{}
}
\email{tingyao.ustc@gmail.com}


\author{Tiejun Zhao}
\orcid{0000-0003-4659-4935}
\authornote{Corresponding Author}
\affiliation{%
 \institution{Harbin Institute of Technology}
 \country{}
}
\email{tjzhao@hit.edu.cn}

\author{Tao Mei}
\orcid{0000-0002-5990-7307}
\affiliation{%
 \institution{HiDream.ai}
 \country{}
}
\email{tmei@live.com}

\renewcommand{\shortauthors}{Mohan Zhou et al.}

\begin{abstract}
A reliable and comprehensive evaluation metric that aligns with manual preference assessments is crucial for conversational head video synthesis methods development. Existing quantitative evaluations  often fail to capture the full complexity of human preference, as they only consider limited evaluation dimensions. Qualitative evaluations and user studies offer a solution but are time-consuming and labor-intensive. This limitation hinders the advancement of conversational head generation algorithms and systems. In this paper, we propose a novel learning-based evaluation metric named \MetricName~(\MetricNameAbbr) for fitting human preference according to the quantitative evaluations across different dimensions. \MetricNameAbbr~can serve as a quantitative evaluation without the need for human annotation. Experimental results validate the superiority of \MetricName~in aligning with human perception, and also demonstrate robustness and generalizability to unseen data, making it a valuable tool for advancing conversation head generation. We expect this metric could facilitate new advances in conversational head generation. Project page: \url{https://github.com/dc3ea9f/PreferenceScore}.
\end{abstract}

\ccsdesc[500]{Computing methodologies~Computer vision problems}

\keywords{evaluation metric; digital human; conversational head generation}

\maketitle

\section{Introduction} \label{sec:intro}
Communication is a ubiquitous activity in people's daily lives, extensively studied by sociologists and linguists over the years. Currently, digital humans are being employed in various roles such as bandwidth-limited video transmission and virtual anchors, serving as brand ambassadors, digital influencers, customer support representatives, and avatars in the Metaverse, among others. The widespread adoption of digital humans enables customers to swiftly access accurate information while enjoying round-the-clock companionship across multiple channels, including desktop, mobile, and tablet platforms, these digital beings accompany them throughout their journey, offering companionship and guidance.

This has led to the emergence of new demands in speech synthesis~\cite{zhou2023visual} and computer vision, \ie, conversational head generation. The objective is to synthesize videos that can actively participate in communication with real human beings. Simultaneously, numerous evaluation metrics are employed to facilitate tangible progress. The capability to precisely assess the performance of generation algorithms and systems serves not only as a quantitative measure of their effectiveness but also empowers researchers to compare various approaches, identify areas for enhancement, and push the boundaries of the current state-of-the-art.

To thoroughly evaluate the quality of synthesized videos, researchers employ various methods, including quantitative evaluation, qualitative evaluation, and user studies. These multifaceted approaches collectively contribute to a thorough understanding of the synthesized videos' performance and effectiveness. However, it is important to note that qualitative evaluation and user studies, while offering valuable insights into user expectations and preferences, often involve manual annotations and cannot be extended to large-scale or high-frequency evaluations. 

\begin{table*}[t]
    \centering
    \footnotesize
    \caption{Bunch of quantitative metrics for conversational head assessment. Metrics selected in our paper are marked with \textbf{bold}.}
    \label{tab:metric}
    \begin{tabular}{lcll}
    \toprule
    Category & Metric & Summarize & References \\
    \hline
    \multirow{7}{*}{Visual quality} & \textbf{SSIM}  & (multi-scale) perceptual similarity for contrast/luminance/structure & \cite{zeng2022fnevr,khakhulin2022realistic,zheng2022avatar,grassal2022neural,liu2022semantic,gafni2021dynamic,yao2022dfa,guan2023stylesync,wang2023seeing,du2023dae,shen2023difftalk,stypulkowski2023diffused,ji2022eamm,liang2022expressive,wang2021audio2head,zhou2021pose,vougioukas2020realistic,chen2020talking,agarwal2023audio,zhou2022responsive,zhou2023interactive} \\
    & \textbf{PSNR} & peak signal-to-noise ratio & \cite{yin2023nerfinvertor,zeng2022fnevr,khakhulin2022realistic,zheng2022avatar,liu2022semantic,gafni2021dynamic,yao2022dfa,guan2023stylesync,wang2023seeing,du2023dae,shen2023difftalk,stypulkowski2023diffused,ji2022eamm,wang2021audio2head,vougioukas2020realistic,agarwal2023audio,zhou2022responsive,zhou2023interactive} \\
    & LPIPS & learned perceptual image patch similarity & \cite{zeng2022fnevr,khakhulin2022realistic,zheng2022avatar,grassal2022neural,gafni2021dynamic,du2023dae,shen2023difftalk} \\
    & \textbf{CPBD} & a perceptual-based no-reference objective image sharpness metric & \cite{grassal2022neural,liu2022semantic,stypulkowski2023diffused,vougioukas2020realistic,zhang2021facial,zhou2022responsive,zhou2023interactive} \\
    & MAE/MSE & mean absolute/square error & \cite{zeng2022fnevr,zheng2022avatar,grassal2022neural,gafni2021dynamic,agarwal2023audio,stypulkowski2023diffused} \\
    & \textbf{FID}  & distance between synthetic and real data distributions & \cite{yin2023nerfinvertor,sun2023next3d,tang2022explicitly,zeng2022fnevr,khakhulin2022realistic,goyal2023emotionally,wang2021audio2head,chen2020talking,prajwal2020lip,agarwal2023audio,zhou2022responsive,zhou2023interactive} \\
    & \textbf{ID} & identity preservation metric such as ArcFace & \cite{yin2023nerfinvertor,zeng2022fnevr,khakhulin2022realistic,grassal2022neural,liu2022semantic,guan2023stylesync,chen2020talking,zhang2021facial,tang2022explicitly,zhou2022responsive,zhou2023interactive} \\
    \hline
    \multirow{4}{*}{Naturalness} & \textbf{ExpL$_n$}  & $L_n$ distance of expression coefficients from some parametric face model  & \cite{sun2023next3d,tang2022explicitly,agarwal2023audio,ng2022learning,zhou2022responsive,zhou2023interactive} \\
     & \textbf{PoseL$_n$} & $L_n$ distance of head pose coefficients from some parametric face model & \cite{sun2023next3d,tang2022explicitly,ng2022learning,zhou2022responsive,zhou2023interactive} \\
     & \textbf{LMD} & landmark distance of mouth/face counter/total & \cite{zeng2022fnevr,zheng2022avatar,liu2022semantic,yao2022dfa,guan2023stylesync,du2023dae,stypulkowski2023diffused,ji2022eamm,liang2022expressive,zhou2021pose,chen2020talking,agarwal2023audio,zhang2021facial,zhou2022responsive,zhou2023interactive} \\
     & Blinks & average eye blinking rate and average inter-blink duration & \cite{yao2022dfa,vougioukas2020realistic,zhang2021facial} \\
    \hline
    \multirow{2}{*}{Speaker-specified} & \textbf{SyncNet} & the synchronization offset and/or confidence of lip motion with audio & \cite{agarwal2023audio,du2023dae,goyal2023emotionally,guan2023stylesync,ji2022eamm,liang2022expressive,liu2022semantic,prajwal2020lip,shen2023difftalk,zhang2021facial,zhou2021pose,zhou2022responsive,zhou2023interactive} \\
    & LipReading & word error rate or character error rate from a lip-reading model & \cite{wang2023seeing,stypulkowski2023diffused,vougioukas2020realistic} \\
    \hline
    \multirow{2}{*}{Listener-specified} & \textbf{ExpFD}  & Fréchet distance in expression coefficients from 3DMM & \cite{ng2022learning,zhou2023interactive} \\
   & \textbf{PoseFD} & Fréchet distance in head pose coefficients from 3DMM & \cite{ng2022learning,zhou2023interactive} \\
    \bottomrule
    \end{tabular}
\end{table*}

On another note, it is worth highlighting the challenge posed by inconsistent quantitative evaluations in the literature. We collected quantitative evaluation metrics that have been used in at least two different works, and present them in~\cref{tab:metric}. By examining the metrics employed across multiple studies, we can observe the persistent endeavors of researchers to align them with human perception. While those bunch of metrics used introduces obstacles for successors to follow and hinder the accurate assessment of synthesized videos as they often deviate from human perception, given their multi-dimensionality. As also demonstrated by the ViCo Challenge 2022 and 2023 leaderboards\footnote{\url{https://vico.solutions/leaderboard/2022}}\footnote{\url{https://vico.solutions/leaderboard/2023}}, it evidenced that the aggregate scores derived from commonly used quantitative metrics, including both the number of Top-1 rankings and the ranking-based scoring system utilizing an additive function, do not consistently align with human preference. This significantly weakens the credibility of quantitative evaluation metrics. It emphasizes the need for more refined and reliable metrics that can align with human preferences.

In this paper, we address the limitations of previous quantitative methods by incorporating human perception into these metrics. We first collect a range of mainstream evaluation methods and design a ranking-based learning approach based on them, utilizing the ranking of manual evaluation results as a guiding principle. Additionally, we propose a video-level data augmentation strategy to construct paired training samples. Through this approach, we train and derive our evaluation metric, \MetricName~(\MetricNameAbbr). We validate the superiority of \MetricNameAbbr~in aligning with human perception on the ViCo 2023 leaderboard. Furthermore, leveraging our learning-based ranking method, we conduct an importance analysis on previous traditional evaluation metrics and identify the most user-engaging attributes for synthesized conversational head videos. We hope that the evaluation metric proposed in this study can effectively guide researchers in efficiently iterating their models, thus contributing to the development of this field.

\section{Related Works} \label{sec:related}
\noindent\textbf{Evaluation metrics for synthesized head videos.}\quad As shown in~\cref{tab:metric}, the quantitative evaluation metrics can be categorized into four distinct dimensions: visual quality, naturalness, speaker-specified, and listener-specified. The inclusion of the last two dimensions is crucial as they pertain to the different roles involved in a conversation, namely the speakers and the listeners. We can observe that different works employ varying comparison methods, making it challenging to achieve fair and consistent performance comparisons among different approaches. In contrast to the established and well-defined evaluation metrics used in traditional computer vision tasks, such as classification,  detection, and segmentation, the evaluation metrics employed in conversational head generation exhibit a heightened level of chaos and uncertainty. The absence of standardized and widely accepted metrics in this domain further amplifies the existing metric chaos, making the evaluation process even more intricate and challenging. To construct a metric that aligns with human preference, we conducted a comprehensive analysis of existing metrics. Considering their relevance and significance, we carefully selected eleven metrics, which are \textbf{boldly} annotated in~\cref{tab:metric}, to form the foundation of our proposed conversational heads evaluation framework.

\noindent\textbf{Human preference evaluation}\quad Deep reinforcement learning refers to a method of training models by incorporating human feedback to guide the learning process. It aims to address the challenge of defining reward functions or objective functions that capture complex human preferences in a given task. The reward model plays a crucial role in human preference evaluation. It can align with humans by capturing relative preferences through pairwise comparisons or rankings. In this paper, we leverage the pairwise training strategy derived from the reward model, building upon the quantitative evaluation results obtained from the ViCo challenge. 

\begin{figure}[h]
  \centering
  \includegraphics[width=1\linewidth]{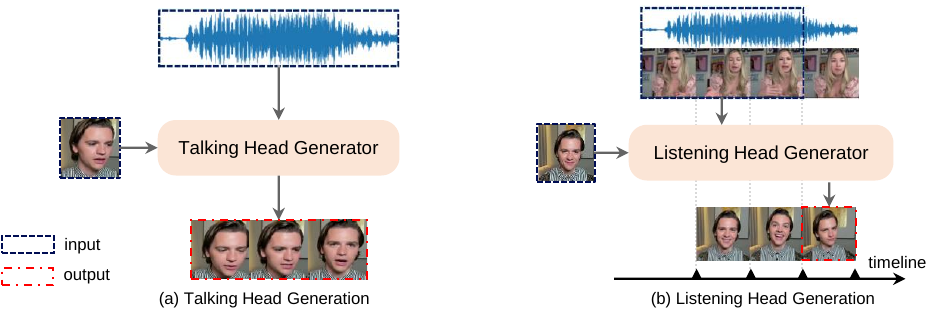}
  \caption{Illustrations of the two tracks in ViCo challenge. (a) Vivid talking head generation, and (b) Responsive listening head video generation.}
  \label{fig:tasks}
\end{figure}

\section{ViCo Challenge} \label{sec:data}
To get annotated human preference data for conversational head generation, we utilize the data from ViCo Challenge. As illustrated in~\cref{fig:tasks}, the challenge includes two tracks: \textbf{Vivid talking head video generation} conditioned on the identity and audio signals of the speaker, \textbf{Responsive listening head video generation} conditioned on the identity of the listener and with responding to the speaker’s behaviors in real-time. The dataset used in this challenge is an extension of the ViCo dataset~\cite{zhou2022responsive}, which consists of conversational video clips obtained from YouTube. The final leaderboard ranking in this challenge is determined using an aggregated scoring function that combines several quantitative evaluations (highlighted in \textbf{bold} of~\cref{tab:metric}). However, it is important to note that this ranking may not fully capture user preferences or satisfaction. To address this limitation, they also design a People's Selection Award, which involves ten experts hailing from production, user experience development, and research domains. These experts are asked to manually rank the challenge submissions and select a team that best satisfies the user. These human preference data are further augmented for constructing the training, validation, and testing set for our proposed metric, as we described in~\cref{sec:augment}.
\section{Method} \label{sec:method}
Our ultimate goal is to develop a metric capable of evaluating human preferences. In a formal sense, we are given a set of $N$ synthesized conversational head videos, denoted as $\mathcal{V}=\{V_1, V_2, \cdots, V_N\}$. The objective is to learn a model $h$ that can assign a human preference score to each video, forming as $\mathcal{S}=\{S_1, S_2, \cdots, S_N\}$, in order to establish a ranking among these synthesized videos.

\subsection{Objective Function}
Since ranking relies on sorting, which is a non-differentiable operation, leading to difficulty in optimization. To address this, we adopt the approach introduced in RankNet~\cite{burges2005learning} by modeling the process with the partially ordered human preference scores $\mathcal{S}$. Specifically, we define the probability of $V_i$ ranking higher than $V_j$ as:
\begin{equation}
    P(V_i \rhd V_j) = \frac{1}{1+e^{-(S_i - S_j)}}.
\end{equation}
To optimize the probability, we utilize the binary cross-entropy loss, which penalizes the discrepancy between the predicted probabilities $P(V_i \rhd V_j)$ and the target ranking $Y_{ij}$:
\begin{equation}
\begin{aligned}
    \mathcal{L} = - \sum_{i, j} Y_{ij}\cdot\log P(V_i \rhd V_j) &+ (1-Y_{ij})\cdot\log(1-P(V_i \rhd V_j), \\
    \mathrm{\textit{where\;}} (i, j) = \{(a, b) &| 1 \leq a < b \leq N, S_a \neq S_b\}.
\end{aligned}
\end{equation}
By optimizing $\mathcal{L}$, the model can generate ranking results of synthesized videos $\mathcal{V}$ from human preference scores $\mathcal{S}$.

Instead of directly modeling the mapping $h$ in video RGB space, we take advantage of existing multidimensional and comprehensive quantitative evaluation metrics. For the video $V_i$, we denote its evaluation results as $M_i=[m_1, m_2, \cdots, m_n]$, where $n$ represents the number of these metrics. This approach significantly reduces the input complexity from $\mathbb{R}^{(|V_i|+|V_j|)\times HWC}$ to $\mathbb{R}^{2\times n}$ during pairwise optimization, making it easier to optimize by avoiding an excessively larger search space. Moreover, using the pre-defined metrics as prior also guarantees that all essential properties are taken into consideration when ranking the videos.

\subsection{Model Architecture}
To mitigate the issue of overfitting, we adopt a set of recommended practices from~\cite{burges2005learning,burges2006learning}. In order to learn the mapping function $h: M\mapsto S$, we employ several fully-connected layers, each equipped with batch normalization and activation functions. Batch normalization aids in stabilizing the training process and allows for faster convergence. It also helps prevent the network from relying too heavily on specific weights or biases, thus reducing the likelihood of overfitting. This architecture allows our model to capture intricate patterns and relationships present in the input data.

Additionally, activation functions are applied after each batch normalization layer. ReLU~\cite{nair2010rectified} is used for the intermediate layers, while ReLU6~\cite{howard2017mobilenets} is used for the final output. These functions introduce non-linearities into the model, enabling it to learn complex mappings between the input data $M$ and the desired output $S$.

\subsection{Data Preparation}\label{sec:augment}
While we can only determine the best video in ViCo Challenge through the People's Selection Award, and the number of participants is limited, the process of human preference score assignment and training data augmentation is indispensable. For each track, we first assign a video human preference score of $10$ if it wins the People's Selection Award; otherwise, the score is set to $0$. Then, we expand the data by intra-track stitched video generation and inter-track fake video expansion. 

\textbf{Intra-track stitched video generation.}\quad A video can be defined as the collection of $k$ clips: $V_i =[v_{i,1}, v_{i, 2}, \cdots, v_{i, k}]$. To generate a stitched video, we utilize the video clips provided by solutions $(t_1, t_2, \cdots t_k) \in A_N^k$, where $A$ is the permutation operation. The resulting stitched video is represented as $[V_{t_1, 1}, V_{t_2, 2}, \cdots, V_{t_k, k}]$. However, due to the stitch operation that disrupts the temporal coherence of the video, the human preference score is set to $10-k$ if the best team is selected; otherwise, it is set to $-k$.

\textbf{Inter-track fake video generation.}\quad After intra-track stitched video generation for each track, we also leverage videos from the other track to provide ``hard negative'' samples. Especially, the other track's videos are injected into the current track with human preference score plus $-100$.

By incorporating these data augmentation techniques, we can get over 500k pairs for training each track. The expanded dataset enables the models to learn from a wider range of examples, including stitched videos and videos from other tracks, ultimately leading to more robust and accurate video assessment models.

\section{Experiment} \label{sec:experiment}
\subsection{Implementation Details}
We select all submissions in ViCo Challenge 2022 for training and validation, then use 2023 data for test. In the ViCo Challenge 2022, the duration of videos varies from 1 to 91 seconds~\cite{zhou2022responsive}. To ensure that the stitched videos maintain a reasonable level of coherence, we set the value of $k$ to be either 2, 3, or 4 for intra-track stitched video generation. After completing the data preparation steps, we proceed to split the dataset into training and validation sets. We allocate 80\% of the prepared data for training purposes, while the remaining 20\% is set aside for validation.

Due to the variation in quantitative evaluation methods between ViCo Challenge 2022 and 2023, we conducted a re-evaluation of the solutions using common metrics. These metrics include SSIM, PSNR, CPBD, FID, ID, ExpL$_1$, PoseL$_1$, as well as specific metrics Lip LMD and SyncNet(AVOffset and AVConf) for vivid talking head generation, plus ExpFD and PoseFD for responsive listening head generation metrics. We employ a separate training approach for each task, as they prioritize distinct aspects within their domains. All metrics are pre-processed with min-max normalization and mean/std standardization before being fed into the model.

\subsection{Performance Comparison}
In order to compare with human preference, we additionally employed ten experts to generate a rank for ViCo Challenge 2023 results. We use the mean reciprocal rank (MRR) and discounted cumulative gain (DCG) to provide insights into the relative rankings and effectiveness of the results.

MRR is the average reciprocal rank of results in a query set. The reciprocal rank is the inverse of the rank of the first correct match. For our task, we evaluate MRR using the reciprocal rank for the query "best video aligning human preference".

DCG is a metric used to evaluate the ranking quality in information retrieval. Mathematically, it can be expressed as:
\begin{equation}
    \mathrm{DCG} = \sum_{i=1}^{N}\frac{rel_i}{\log_2(i+1)},
\end{equation}
where $rel_{i}$ is the graded relevance of the result at position $i$. It offers an assessment of the ranking list's overall quality, providing a broader perspective compared to MRR.

To thoroughly evaluate the performance, we compare our model with naive ranking methods and machine learning methods:
\begin{enumerate}
    \item \textbf{Num of Top-1s (\#Top1)} is the ranking method used in ViCo Challenge 2022, representing the instances where a solution achieved the top rank. Participants have expressed concerns about its bias towards models excelling in specific metrics while underperforming in others.
    \item \textbf{Ranking-based Scoring (RS)} is the ranking method used in ViCo challenge 2023. This scoring system takes into account the overall ranking of the participating methods or models, and uses the additive function for its rank, providing a comprehensive evaluation of their performance.
    \item \textbf{Decision Tree (DT)} is a popular machine learning algorithm known for its versatility and interpretability. Here, we utilize a decision tree classification model to enhance the model's awareness of varying levels of human preferences.
    \item \textbf{CatBoost (CB)}~\cite{prokhorenkova2018catboost, dorogush2018catboost} is a gradient-boosting algorithm that leverages gradient-boosting techniques to construct an ensemble of decision trees, leading to improved predictive accuracy. Similar to the decision tree, we use the CatBoost classifier to train a scoring model.
\end{enumerate}

The evaluation results are presented in~\cref{tab:performance,tab:leaderboard}. It is evident that the hand-crafted metrics, specifically \#Top-1 and RS, exhibit poorer performance when compared to machine learning (DT, CB) and deep learning techniques (PS). DT is overfitting to training data and may be too easy to fully capture the complexity of given quantitative metrics, resulting in suboptimal performance.

Our deep learning approach outperforms both hand-craft methods and machine learning methods in terms of MRR and DCG, especially for listening heads. The superior results indicate that our model indeed learns and evaluates human preference. Furthermore, the evaluation conducted on ViCo 2023 showcases the robustness and generalizability of our approach. 

\begin{table}[t]
\centering
\caption{Performance comparison among different methods. ``GT'' indicates the best value when the rank perfectly matches human preference. ``DT'', ``CB'' and ``PS'' were executed five times to calculate the mean and std for MRR and DCG.}
\label{tab:performance}
\resizebox{\columnwidth}{!}{
\begin{tabular}{cccccc}
\toprule
    \multirow{2.5}{*}{Method} & \multicolumn{2}{c}{Talking Heads} & \phantom{a} & \multicolumn{2}{c}{Listening Heads} \\
    \cmidrule{2-3} \cmidrule{5-6}
    & MRR & DCG && MRR & DCG \\
\midrule
    GT & $1$ & $10.27$ && $1$ & $13.58$ \\
\hdashline[.4pt/1pt]
    \#Top1 & $1$ & $9.65$ && $0.25$ & $11.80$ \\
    RS     & $1$ & $9.75$ && $0.25$ & $11.87$ \\
\hdashline[.4pt/1pt]
    DT     & $0.452\pm0.285$ & $8.815\pm0.799$ && $0.415\pm0.287$ & $11.635\pm1.016$ \\
    CB     & $1.000\pm0.000$ & $9.925\pm0.223$ && $0.409\pm0.256$ & $11.876\pm0.863$ \\
   \MetricNameAbbr    & $0.929\pm0.175$ & $10.075\pm0.151$ && $0.917\pm0.186$ & $12.527\pm0.397$ \\
\bottomrule
\end{tabular}
}
\end{table}

\begin{table}[t]
\centering
\caption{Ranking score comparison of top solutions. 1--3 lines for Talking Heads results and 4--7 lines for Listening Head results. $^*$ indicates the People's Selection Award.}
\label{tab:leaderboard}
\resizebox{\columnwidth}{!}{
\begin{tabular}{lcccccc}
\toprule
Solution & \#Top-1$\uparrow$ & RS$\downarrow$ & DT$\uparrow$ & CB$\uparrow$ & PS$\uparrow$ \\
\midrule
ilearn$^*$        & 5 & 12 & $4.00\pm1.67$ & $6.00\pm0.00$ & $4.33\pm0.55$ \\
Robo Space        & 4 & 13 & $4.00\pm1.67$ & $4.40\pm1.20$ & $2.87\pm0.72$ \\
metah             & 0 & 22 & $4.00\pm1.67$ & $3.20\pm1.47$ & $0.24\pm0.29$ \\
\hdashline[.4pt/1pt]
DXM-DI-AI-CV-TEAM & 5 &  7 & $1.40\pm0.80$ & $1.00\pm0.00$ & $3.49\pm0.46$ \\
Robo Space        & 3 &  9 & $1.00\pm0.89$ & $0.20\pm0.40$ & $0.00\pm0.00$ \\
ilearn            & 1 & 15 & $1.00\pm0.89$ & $0.60\pm0.49$ & $4.23\pm0.60$ \\
Lyalunov$^*$      & 0 & 36 & $1.60\pm0.49$ & $1.00\pm0.00$ & $6.00\pm0.00$ \\
\bottomrule
\end{tabular}
}
\end{table}

\subsection{Feature Importance Analysis}
We utilize permutation feature importance (PI), a valuable technique for enhancing the interpretability of black-box deep learning methods. This approach quantifies the reduction in model performance when a single feature value is randomly shuffled, thus indicating the feature's significance. Given that the present Mean MRR value primarily focuses on the top-ranked video and its variance can be fluctuating when features are permuted, we opt for DCG to assess the relative importance of features in relation to human preference.

We have observed that for talking heads, the AVConf feature holds the highest influence, leading to a decrease in DCG from 10.20 to 8.05. Surprisingly, this performance is even inferior to that achieved by hand-crafted methods. Conversely, in the case of listening heads, the most influential feature is PoseFD, resulting in a decrease in DCG from 13.02 to 11.91. This observation aligns with our cognitive priorities when evaluating synthesized videos. When it comes to talking heads, people tend to pay close attention to the synchronization of mouth movements with the audio. Similarly, for listening heads, people prioritize assessing whether their behavior resembles that of a typical listener, as indicated by PoseFD.
\section{Conclusion} \label{sec:conclusion}
In conclusion, this paper has addressed the pressing need for human preference evaluation metrics in the field of conversation head generation. We have introduced the \MetricName~(\MetricNameAbbr) metric, which successfully overcomes the limitations of existing chaotic quantitative evaluation metrics. Through experiments, we have validated the superiority of our metric in aligning with human perception, demonstrating its robustness and generalizability even with unseen data. We firmly believe that the \MetricName~will prove its valuable in advancing conversational head generation. This advancement holds immense potential for driving applications in a wide range of scenarios, thereby opening up new avenues for development and innovation.


\bibliographystyle{ACM-Reference-Format}
\balance
\bibliography{sample-base}
\end{document}